\documentclass[graphics]{llncs}

\usepackage{amsmath}
\usepackage{amssymb}
\usepackage{epsfig}

\begin{document}

\title{LSIS Research Report 2003-06-004 \\Pruning Isomorphic Structural Sub-problems in Configuration}

\author{Stephane Grandcolas and Laurent Henocque and Nicolas Prcovic}

\institute{Laboratoire des Sciences de l'Information et des Syst\`emes\\
LSIS (UMR CNRS 6168)\\
Campus Scientifique de Saint J\'er\^ome\\
Avenue Escadrille Normandie Niemen\\
13397 MARSEILLE Cedex 20}

\maketitle

\begin{abstract}
Configuring consists in simulating the realization of a complex product from a catalog of component parts, using known
relations between types, and picking values for object attributes. This highly combinatorial problem in the field of
constraint programming has been addressed with a variety of approaches since the foundation system R1\cite{McDermott82}. An
inherent difficulty  in solving configuration problems is the existence of many isomorphisms among interpretations. We
describe a formalism independent  approach to improve the detection of isomorphisms by configurators, which does not
require to adapt the problem model. To achieve this, we exploit the properties of a characteristic subset of configuration
problems, called the structural sub-problem, which canonical solutions can be produced or tested at a limited cost. In this
paper we present an algorithm for testing the canonicity of configurations, that can be added as a symmetry breaking
constraint to any configurator. The cost and efficiency of this canonicity test are given.
\end{abstract}


\section{Introduction}
Configuring consists in simulating the realization of a complex product from a catalog of component parts (e.g. processors,
hard disks in a PC ), using known relations between types (motherboards can connect up to four processors), and
instantiating object attributes (selecting the ram size, bus speed, \ldots). Constraints apply to configuration problems to
define which products are valid, or well formed. For example in a PC, the processors on a motherboard all have the same
type, the ram units have the same wait times, the total power of a power supply must exceed the total power demand of all
the devices. Configuration applications deal with such constraints, that bind variables occurring in the form of variable
object attributes deep within the object structure.

The industrial need for configuration applications is widespread, and has triggered the development of many configuration
applications, as well as generic configuration tools or configurators, built upon all available technologies. For instance,
configuration is a leading application field for rule based expert systems. As an evolution of R1\cite{McDermott82}, the
XCON system \cite{barkerconnor89} designed in 1989 for computer configuration at Digital Equipment involved  31000
components, and 17000 rules. The application of configuration is experimented or planned in many different industrial
fields, electronic commerce (the CAWICOMS project\cite{Felfernig02}), software\cite{Ylinen02}, computers\cite{Plain2002},
electric engine power supplies\cite{JohnGeske99} and many others like vehicles, electronic devices, customer relation
management (CRM) etc.


The high variability rate of configuration knowledge (parts catalogs may vary by up to a third each year) makes
configuration application maintenance a challenging task. Rule based systems like R1 or XCON lack modularity in that
respect, which encouraged researchers to use variants of the CSP formalism (like DCSP
\cite{Mittal90AA,soininen99fixpoint,AFM02}, structural CSP \cite{Nareyek99}, composite CSP
\cite{sabinfreudercomposite96}), constraint logic programming (CLP \cite{JL87}, CC \cite{FSB99}, stable models
\cite{soininen01representing}), or object oriented approaches\cite{Mailharro98,Meyer99}. 


One difficulty with configuration problems stems from the existence of many isomorphisms among interpretations.
Isomorphisms naturally arise from the fact that many constraints are universally quantified (e.g. "\emph{for all
motherboards, it holds that their connected processors have the exact same type}"). This issue is technically discussed in
several papers\cite{Mailharro98,Weigel1998,Tiihonen02}. The most straightforward approach is to treat during the search
all yet unused objects as interchangeable. This is a widely known technique in constraint programming, applied to
configuration in  \cite{Mailharro98,Tiihonen02} e.g.. However, this does not account for the isomorphisms arising during
the search because substructures are themselves isomorphic (e.g. two exactly identical PCs with the same motherboards and
processors are interchangeable).

The work in \cite{Mailharro98}, implemented within the ILOG\footnote{http://www.ilog.fr} commercial configurators,
suggests to replace some relations between objects with cardinality variables counting the number of connected elements for
each type. This technique is very efficient and intuitively addresses many situations. For instance, to model a purse, it
suffices to count how many coins of each type it contains, and it would be lost effort to model each coin as an isolated
object. This solution has two drawbacks : it requires a change in the model on one hand, and the counted objects cannot
themselves be configured. Hence the isomorphisms arising from the existence of isomorphic substructures cannot be handled
this way.

\cite{Weigel1998} applies a notion called "context dependant interchangeability" to configuration. This is more general
than the two approaches seen before, but applies to the specific area of case adaptation. Also, since context dependant
interchangeability detection is non polynomial, \cite{Weigel1998} only involves an approximation of the general concept.
Furthermore, the underlying formalism, standard CSPs, is known as too restrictive for configuration in general.

One step towards dealing with the isomorphisms emerging from structural equivalence in configurations is to isolate this
"structure", and study its isomorphisms. This is the main goal pursued here : we propose a general approach for the
elimination of structural isomorphisms in configuration problems. This generalizes already known methods (the
interchangeability of "unused" objects, as well as the use of cardinality counters) while not requiring to adapt the
configuration model. After describing what we call a configurations's \emph{structural sub-problem}, we define an algorithm
to test the canonicity of its interpretations. This algorithm can be adapted to complement virtually any general purpose
configuration tool, so as to prevent exploring many redundant search sub-spaces. This work greatly extends the
possibilities of dealing with configuration isomorphisms, since it does not require a specific formalism. The complexity of
the canonicity test and the compared complexity of the original problem versus the resulting version exploiting canonicity
testing are studied. 

The paper is structured as follows : section 2 describes configuration problems, and the formalism used throughout the
paper. Section 3 defines structural sub-problems, and their models called T-trees. In section 4, we describe T-tree
isomorphisms and their canonical representatives. Section 5 presents an algorithm to test the canonicity of T-trees. Then
section 6 lists complexity and combinatorial results. Finally, 6 concludes and opens various perspectives.

\section{Configuration problems, and structural sub-problems}

A configuration problem describes a generic product, in the form of declarative statements (rules or axioms) about product
well-formedness.  Valid configuration model instances are called \emph{configurations}, are generally numerous, and involve
objects and their relationships. There exist several kinds of relations :
\begin{itemize}
\item \emph{types} : unary relations involved in taxonomies, with inheritance. They are central to configuration problems
since part of the objective is to determine, or refine, the actual type of all objects present in the result (e.g. : the
program starts with something known as a "Processor", and the user expects to obtain something like
"Proc\_\emph{Brand}\_\emph{Speed}").
\item other unary relations corresponding to Boolean object properties (e.g. : a main board has a built in scsi interface)
\item binary \emph{composition} relations (e.g. : car wheels, the processor in a mainboard \ldots). An object cannot act as
a component for more than one composite.
\item other relations : not necessarily  binary, allowing for loose connections (e.g. : in a computer network, the relation
between computers and printers)
\end{itemize}
Configuration problems generally exhibit solutions having a prominent structural component, due to the presence of many
composition relations. Many isomorphisms exist among the structural part of the solutions. 
We isolate configuration sub-problems called \emph{structural problems}, that are built from the composition relations, the
related types and the structural constraints alone. By \emph{structural constraints}, we precisely refer to the basic
constraints that define the structure : 
\begin{itemize}
\item those declaring the types of the objects connected by each relation
\item the constraints that specify the maximal cardinalities of the relations (the maximal number of connectable
components)
\end{itemize}
To ensure the completeness of several results at the end of the paper, we enforce two limitations to the kind of
constraints that define structural problems : minimal cardinality constraints are not accounted for at that level (they
remain in the global configuration model), and the target relation types are all mutually exclusive\footnote{this can be
compensated for by using zero max cardinality constraints in the global configuration problem}. 

For simplicity, we abstract from any configuration formalism, and consider a totally ordered set $O$ of objects (we
normally use $O=\{1,2,\ldots \}$), a totally ordered set $T_C$ of type symbols (unary relations) and a totally ordered set
$R_C$ of composition relation symbols (binary relations). We note $\prec_{O}$, $\prec_{T_C}$ and $\prec_{R_C}$ the
corresponding total orders.

\begin{definition}[syntax]
A \emph{structural problem}, as illustrated in figure \ref{structural_example}, is a tuple $(t,T_C,R_C,C)$, where $t \in
T_C$ is the root configuration type, and $C$ is a set of structural constraints applied to the elements of $T_C$ and $R_C$.
\end{definition}

\begin{figure}[htb]
\vspace{-0.3cm}
\begin{center}
\begin{tabular}{|@{\hspace{4mm}}p{9cm}@{\hspace{4mm}}|}
\hline \vspace{0.0cm}
$t$ = PC\\
$T_C$ = \{PC, Monitor, Supply, Mainboard, Processor, HDisk\}\\
$R_C$ = \{PC-Monitor, PC-Supply, PC-Mainboard, Mainboard-Processor, Mainboard-HDisk\} \\
$C$ =  \{
\hspace{0.0cm}   $\forall \rm{x,y}\ $ PC-Monitor(x,y) $\rightarrow$  PC(x) $\wedge$ Monitor(y), \ldots\\
\hspace*{1cm}   $\forall \rm{x}\ \mid \{y\ st.\ PC\_Monitor(x,y)\}\mid < 2$, \ldots\\
\hspace*{1cm}   $\forall \rm{x}\ \rm{PC(x)}\ \rightarrow \neg \rm{Monitor(x)}$, \ldots \}
\\
\hline
\end{tabular}
\end{center}
\caption{Structural problem example}
\label{structural_example}
\vspace{-0.5cm}
\end{figure}

\begin{definition}[semantics]
An \emph{instance} of a structural problem $(t,T_C,R_C,C)$ is an interpretation $I$ of $\ t$ and of the elements of $T_C$
and $R_C$, over the set $O$ of objects. 
If an interpretation satisfies the constraints in $C$, it is a \emph{solution} (or \emph{model}) of the structural problem.
\end{definition}
In the spirit of usual finite model semantics, $T_C$ members are interpreted by elements of $\mathcal{P}(O)$, and $R_C$
members by elements of $\mathcal{P}(O \times O)$ (relations). For instance, an interpretation of the type "Processor" can
be \{4,6\}, which means that 4 and 6 alone are processors. Similarly, an interpretation of the binary relation
"Mainboard-Processor" can be \{(1,4),(2,6)\}.

For readability reasons and unless ambiguous, in the rest of the paper we use the term \emph{configuration} to denote a
model of a structural problem. Figure \ref{example_configuration} lists a sample model of the structural problem detailed
in figure \ref{structural_example}. It is obvious from this example that object types can be inferred from the composition
relations. We define the following :

\begin{definition}[root, composite, component]
A \emph{configuration}, solution of a structural problem $(t,T_C,R_C,C)$, can be described by the set $U$ of
interpretations of all the elements of $R_C$. If $R_U$ denotes the union of the relations in $U$ ($R_U = \bigcup_{rel \in
U}rel$), and $R_t$ denotes its transitive closure, then we have :
\begin{enumerate}
\item $\exists !\ root\in O$ called {\em root of the configuration}\footnote{root unicity does not restrict generality,
since this can be achieved if needed by introducing an extra type and an extra relation.} for which
$\forall o\in O\ (o,root)\not\in R_U$, 
\item $\forall o \in O$ s.t. $o \neq root$, $\exists !\ c \in O$ s.t. $(c,o)\in R_U$ ; \\we call $c$ the {\em composite} of
$o$ and $o$ a {\em component} of $c$,
\item $\forall o \in O$ s.t. $o \neq root$, $(root,o)\in R_t$.
\end{enumerate}
\end{definition}
Figure \ref{example_configuration} lists a configuration of the problem described in figure \ref{structural_example}.
\begin{figure}[htb]
\begin{center}
\begin{tabular}{|@{\hspace{4mm}}p{9cm}@{\hspace{4mm}}|}
\hline \\
$I$(PC-Monitor) = \{(1,2)\},\\
$I$(PC-Supply) = \{(1,3)\},\\
$I$(PC-Mainboard) = \{(1,4)\},\\
$I$(Mainboard-Processor) = \{(4,5),(4,6)\},\\
$I$(Mainboard-HDisk) = \{(4,7),(4,8)\}\\
$I$(PC) = \{1\}, \ldots $I$(HDisk) = \{7,8\},\\
\\
\hline
\end{tabular}
\end{center}
\caption{A solution of the structural problem of the figure \ref{structural_example}}
\label{example_configuration}
\end{figure}
\vspace{-1cm}
\section{Isomorphisms}
From a  practical standpoint, as soon as two objects of the same type appearing in a configuration are interchangeable, it
is pointless to produce all the  isomorphic solutions obtained by exchanging them. Two solutions that differ only by the
permutation of interchangeable objects are redundant, and the second has no interest for the user. It would be particularly
useful for a configurator to generate only one representative of each equivalence class.
More interestingly, the capacity of skipping redundant interpretations also prunes the search space from many sub-spaces,
and was shown a key issue in other areas of finite model search \cite{AudHen01}.

\begin{definition}
We note $U(rel)$ the relation interpreting the relational symbol  $rel \in R_C$ in $U$. Two configurations $U$ and $U'$ are
\emph{isomorphic} if and only if there exists a permutation $\theta$ over the set $O$, such that $\forall r \in R_c,
\theta(U)(r) = U'(r)$
\end{definition}

\subsection{Coding configurations, T-trees}
\begin{figure}[htb]
  \begin{center}
      \epsfig{file=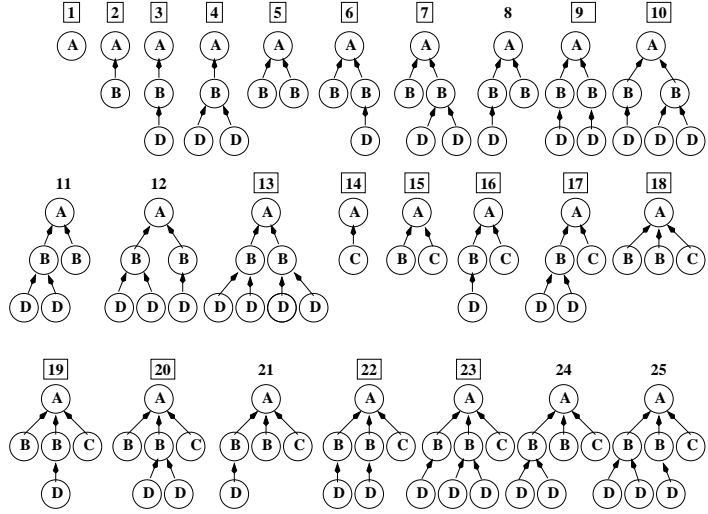,height=7cm}
    \caption{The first 26 T-trees ordered by $\curlyeqprec$, for a problem where at most two objects of type D can connect
to an object of type B, and two objects of types B and C may connect to an object of type A. The numbers of the
$\curlyeqprec$-minimal representatives are framed. 
}
    \label{fig:enumeration}
  \end{center}
\vspace{-0.8cm}
\end{figure}

Because composition relations bind component objects to at most one composite object, configurations can naturally be
represented by trees. For practical reasons, we make the hypothesis that two distinct relations cannot share both their
component and composite types\footnote{without loss of generality : a composition relation can be replaced by two
composition relations plus a new extra type}. 
Then any configuration $U$ is in one to one correspondence with an ordered tree where : 
\begin{enumerate}
\item nodes are labeled by objects of $O$,
\item edges are labeled by the component side type of the corresponding relation,
\item child nodes are sorted first by their type according to $\prec_{T_C}$, then by their label according to $\prec_O$.
\end{enumerate} 
\begin{figure}[htb]
\vspace{-0.3cm}
  \begin{center}
\begin{tabular}{|@{\hspace{4mm}}p{9cm}@{\hspace{4mm}}|}
\hline \\
      \epsfig{file=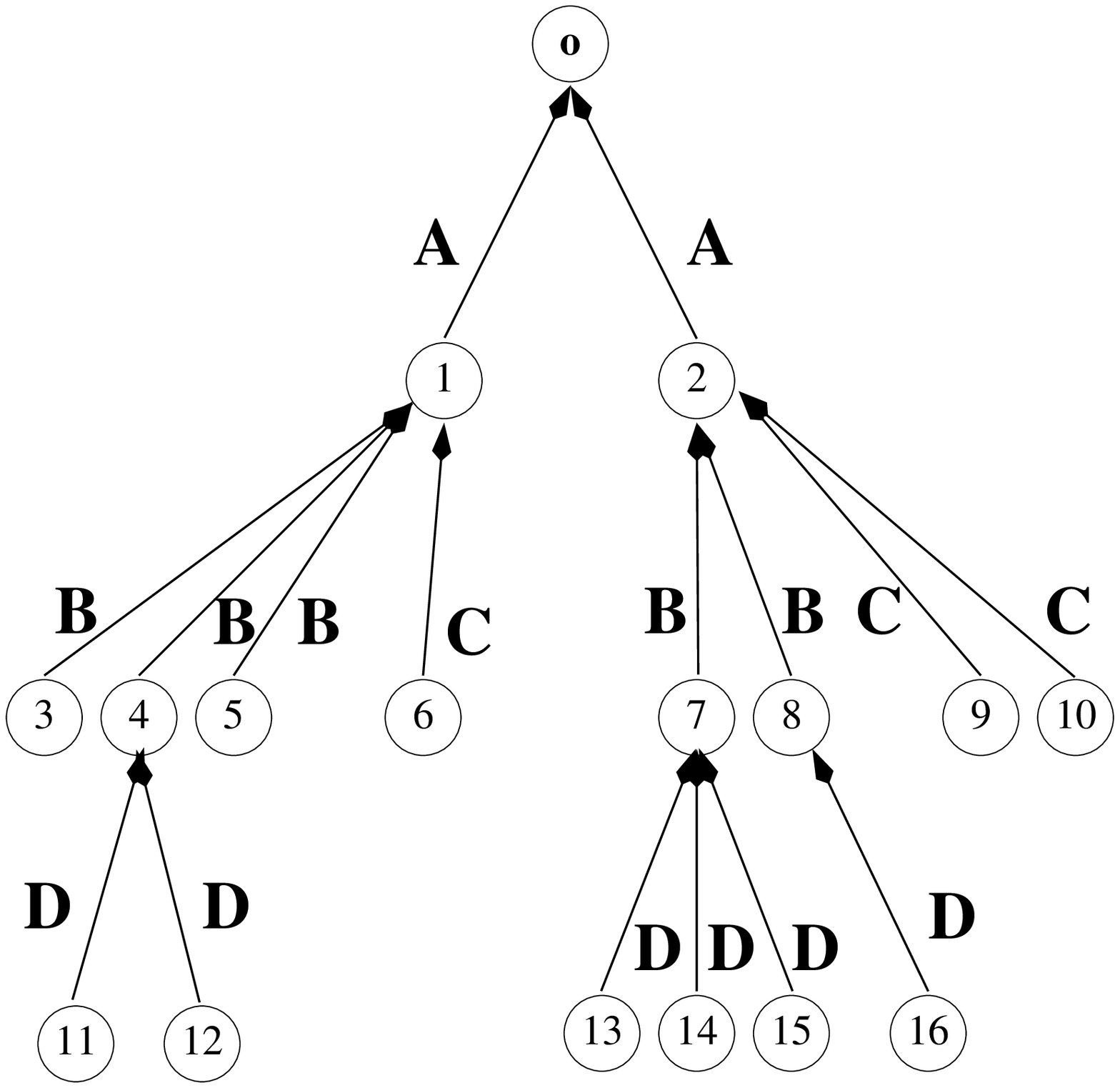,height=4cm}
      \epsfig{file=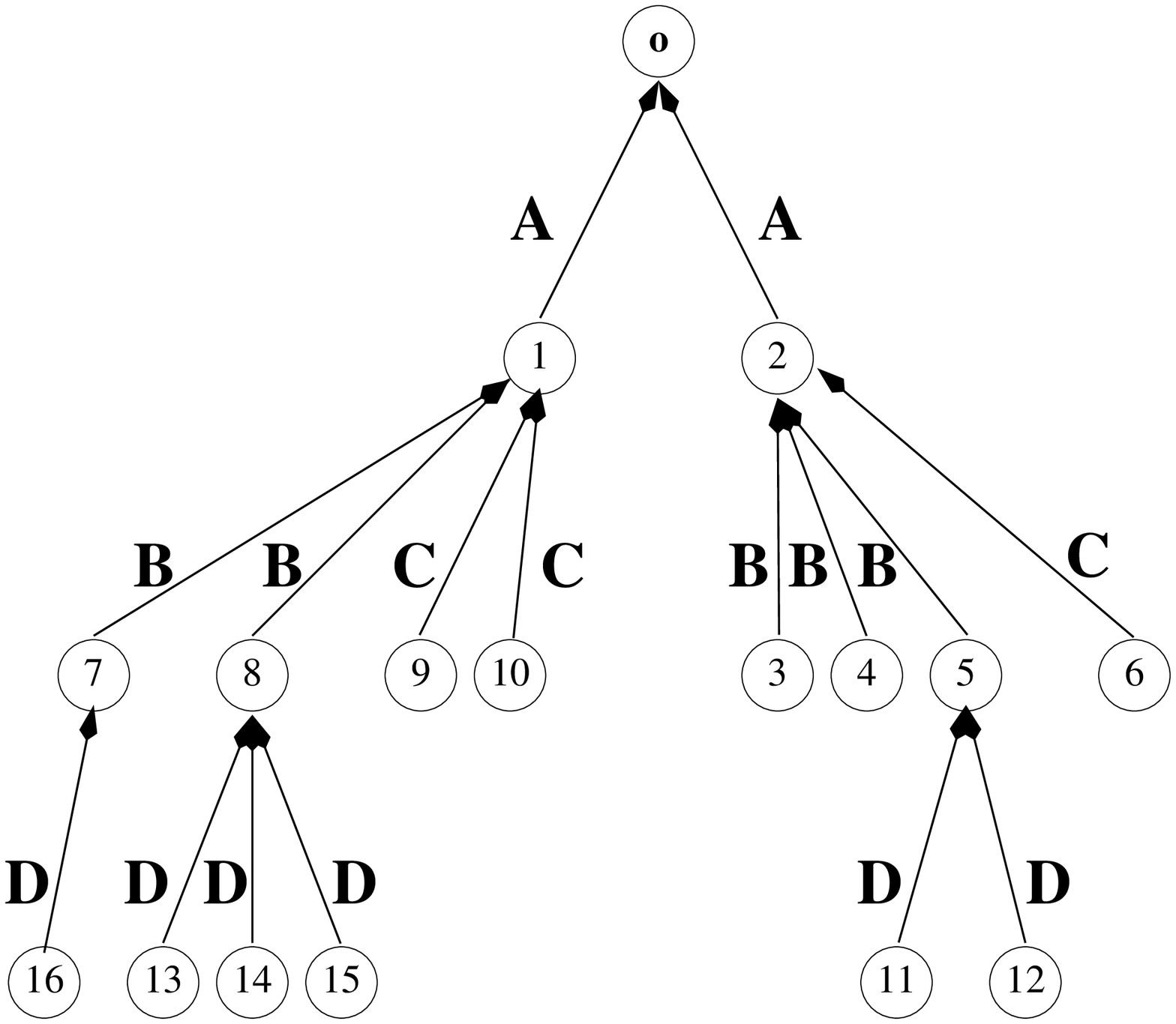,height=4cm}
      \\
\hline
\end{tabular}
  \end{center}
    \caption{Two isomorphic configuration trees.}
    \label{fig:config-tree}
\vspace{-0.3cm}
\end{figure}
Figure \ref{fig:config-tree} illustrates this translation by an artificial example, which shows that object numbers are redundant. If we suppress them, we keep the possibility to
produce a configuration tree isomorphic to the original via a breadth first traversal. We hence introduce \emph{T-trees},
which capture part of the isomorphisms that exist among configurations :
\begin{definition}[T-tree]
A {\em T-tree} is a finite and non empty ordered tree where nodes are labeled by types and children are ordered according
to $\prec_{T_C}$.
We note $(T,\left<c_1,\ldots c_k\right>)$ the T-tree with sub-trees $c_1,\ldots c_k$ and root label $T$.
\end{definition}
To translate a configuration tree in a T-tree, we simply replace the node labels by their parent edge labels. Several
T-tree examples are listed by the figure \ref{fig:enumeration}. To perform the opposite operation, i.e. build a
configuration tree from a T-tree, it suffices to generate node labels via a breadth first traversal (using consecutive
integers, the root being labeled $0$), then to relabel the edges.
\begin{proposition}
Let $A_1$ be a configuration tree, $C_1$ the corresponding T-tree , and $A_2$ the configuration tree rebuilt from $C_1$.
Then $A_1$ and $A_2$ are isomorphic. 
\end{proposition}
The proof is straightforward. A permutation $\theta :\ O \mapsto O$ which asserts the isomorphism can be built by simply
superposing $A_1$ and $A_2$. Since every configuration bijectively maps to a configuration tree, this result legitimates
the use of T-trees to represent configurations. This encoding captures many isomorphisms, because the references to members
of the set $O$ are removed, and the children ordering respects $\prec_{T_C}$.

\subsection{A total order over T-trees}
Configuration trees and T-trees being trees, they are isomorphic, equal, superposable, under the same assumptions as
standard trees. 

\begin{definition}[Isomorphic T-trees ]
Let $C=(T,\left<a_1,\ldots,a_k\right>)$ and $C'=(T',\left<b_1,\ldots,b_l\right>)$ be two T-trees.\\
{\bf Isomorphism :}
$C$ and $C'$ are \emph{isomorphic} ($C \equiv C'$) if $T=T'$, $k=l$ and there exists a bijection $\sigma :
\{a_1,\ldots,a_k\} \mapsto \{b_1,\ldots b_k\}$ such that $\forall i\ \sigma(a_i) \equiv b_i$. $Iso(C)$ denotes the set of
trees which are isomorphic to a T-tree $C$.\\
{\bf Equality :} 
$C$ and $C'$ are \emph{equal} ($C=C'$) if $k=l$, $T=T'$, and $\forall i\ a_i = b_i$.\\
\end{definition}

\begin{proposition}
Two configurations are isomorphic iff their corresponding T-trees are.
\end{proposition}
As a means of isolating a canonical representative of each equivalence class of T-trees, we define a total order over
T-trees.
We note $nct(T)$ (\underline{n}umber of \underline{c}omponent \underline{t}ypes) the number of types $T_i$ having $T$ as
composite type for a relation in $R_C$.
The types $T_i$ ($1 \leq i \leq nct(T)$) are numbered on each node according to $\prec_{T_C}$.
If C is a T-tree, we call {\em T-list} and we note $T_i(C)$ the list of its children having $T_i$ as a root label.
$|T_i(C)|$ is the number of T-trees of the T-list $T_i(C)$.
To simplify list expressions in the sequel,
we use $\left< a_i \right>_{1}^{n}$ to denote the list $\left<a_1, a_2, ..., a_n\right>$.
Many ways exist to recursively compare trees, by using combined criteria (root label, children count, node count, etc.).
For rigor, we propose a definition using two orders $\curlyeqprec$ and $\ll$.

\begin{definition}[The relations $\curlyeqprec$, $\curlyeqprec_{lex}$, $\ll$ and $\ll_{lex}$]\\
We define the following four relations : $\curlyeqprec$
compares T-trees with roots of the same type $T$,
$\curlyeqprec_{lex}$ is its lexicographic generalization to T-lists,
$\ll$ compares two T-lists of same type $T_i$, and $\ll_{lex}$ is its lexicographic generalization to lists $\left< T_i(C)
\right>_{1}^{nct(T)}$.
These four order relations recursively define as follows :
\begin{itemize}
\item $\forall T \in T_C\ :\ (T,\left< \right>) \curlyeqprec (T,\left< \right>)$.
\item $\forall C,\ C'\ \neq (T,\left< \right>)$ : $C \curlyeqprec C'
\iff \left< T_i(C) \right>_{1}^{nct(T)} \ll_{lex} \left< T_i(C') \right>_{1}^{nct(T)}$.
\item $\forall C,\ C'\ \neq (T,\left< \right>)$, $\forall i$ : $T_i(C) \ll T_i(C') \iff \\ |T_i(C)| < |T_i(C')| \vee
|T_i(C)| = |T_i(C')| \wedge T_i(C) \curlyeqprec_{lex} T_i(C')$.
\end{itemize}
\end{definition}
In other words, each T-tree is seen as if built from a root of type $T$ and a list of T-lists of sub-trees. These two list
levels justify having two lexicographic orders. $\curlyeqprec$ (lines 1 and 2) lexicographically compares the lists of
T-lists of two trees having the same root type. $\ll$ lexicographically compares T-lists (taking their length into
account).

\begin{proposition}
The relations $\curlyeqprec$, $\curlyeqprec_{lex}$, $\ll$ and $\ll_{lex}$ are total orders.
\end{proposition}
\begin{proof}
As any lexicographic order defined from a total order is itself total, it remains to prove that the relations
$\curlyeqprec$ and $\ll$ are total orders.
To demonstrate that a binary relation is a total order it suffices to show that any two elements from the set of reference
can be compared, either one being less than or equal to the other. The proof is by induction on the height of T-trees.
\begin{itemize}
\item there exists only one T-tree of height 0 having a root labeled with the type $T$ : $(T, \left< \right>)$. $\forall
T$, $(T, \left< \right>) \curlyeqprec (T, \left< \right>)$.
\item assume that for any two T-trees $C$ and $C'$ of height less than $h$, either $C \curlyeqprec C'$ or $C' \curlyeqprec
C$ holds.
Any couple of T-lists $L = \left<c_1, ... c_{|L|}\right>$ and $L' = \left<c_1', ... c_{|L' |}'\right>$  with height $h+1$
(containing T-trees one of which at least is of height $h$) is such that :
\begin{itemize}
\item if $L = L' $  then $L \ll L' $ (and as well $L'  \ll L$)
\item else $|L| \neq |L'|$ and hence either $L \ll L'$ or $L' \ll L$
\item else $|L| = |L'|$ and then $\exists j$, $\forall i < j$, $c_i = c_i'$ and either $c_j \curlyeqprec c_j'$ or $c_j'
\curlyeqprec c_j$.
Either $L \curlyeqprec_{lex} L' $ or $L' \curlyeqprec_{lex} L$, hence either $L \ll L'$ or $L' \ll L$
\end{itemize}
In all cases, $L \ll L'$ or $L'  \ll L$.
\item now assume that any couple of T-lists $L$ and $L'$ which T-trees have height less than $h$ is such that either $L \ll
L'$ or $L' \ll L$.
Any couple of T-trees $C = (T, \left<l_1, ... l_{nct(T)}\right>)$ and
$C' = (T, \left<l_1', ... l_{nct(T)'}\right>)$ of height $h$ is such that :
\begin{itemize}
\item if $C = C'$  then $C \curlyeqprec C'$ (and as well $C' \curlyeqprec C$).
\item else $\exists j$, $\forall i < j$, $l_i = l_i'$ and either $l_j \ll l_j'$ or $l_j' \ll l_j$.
As a consequence, either $C \ll_{lex} C'$ or $C' \ll_{lex} C$ hence either $C \curlyeqprec C'$ or $C' \curlyeqprec C$.
\end{itemize}
In all cases, $C \curlyeqprec C'$ or $C' \curlyeqprec C$.
\end{itemize}
We call $P(h)$ the property ``\emph{any couple of T-trees $C$ and $C'$ of heigh less than $h$ is such that $C \curlyeqprec
C'$ or $C' \curlyeqprec C\ $}''
and $Q(h)$ the property ``\emph{any couple of T-lists $L$ and $L'$ which T-trees are of height less than $h$ is such that
$L \ll L'$ or $L' \ll L\ $ }''.
We have shown that $P(0)$ is true, and that $\forall h$, $P(h)$ implies $Q(h)$ and $\forall h$, $Q(h)$ implies $P(h+1)$. We
conclude that $\forall h$, $P(h)$ and $Q(h)$, and hence that the relations $\curlyeqprec$ and $\ll$ are total orders, as
are their lexicographic extensions.

\end{proof}

\begin{definition}[Canonicity  of a T-tree]
A T-tree $C$ is {\em canonical} iff it has no child or if $\forall i$, $T_i(C)$ is sorted by $\curlyeqprec$ and $\forall c
\in T_i(C)$, $c$ itself is canonical.
\end{definition}

\begin{proposition}\label{canonicity}
A T-tree is the  $\curlyeqprec$-minimal representative of its equivalence class (wrt. T-tree isomorphism) iff it is
canonical.
\end{proposition}

\begin{proof}
Let C and C' be two isomorphic and distinct T-trees. Consider the following prefix recursive traversal  of a T-tree :
\begin{itemize}
\item examining a T-tree C, is examining its lists $T_i(C)$ in sequence.
\item examining a list $T_i(C)$, is examining its length then, if the length is non zero, examining its T-trees in
sequence.
\end{itemize}
$\Leftarrow$ We first show by induction that if, according to this traversal, two trees differ somewhere by the length of
two T-lists, they are comparable accordingly. 
Compare C and C' by performing a simultaneous prefix traversal, and stop as soon as we meet at depth $p$ two lists
$T_i(S_n)$ and $T_i(S_n')$ with distinct lengths, $S_n$ (resp. $S_n'$) being a sub-tree in C (resp. C').
Call $S$ (resp. $S'$) the parent T-tree of $S_n$ (resp. $S_n'$).
Suppose that $|T_i(S_n)| < |T_i(S_n')|$. 
It follows that $T_i(S_n) \ll T_i(S_n')$. 
Since $\forall j < i, T_j(S_n) = T_j(S_n')$, we have $\left< T_j(S_n) \right>_{1}^{|S_n|} \ll_{lex} \left< T_j(S_n')
\right>_{1}^{|S_n'|}$
and hence $S_n \curlyeqprec S_n'$. Similarly, as $\forall j<n$, $S_j = S_j'$
it follows $L = \left< S_j \right>_{1}^{nct(T)} \curlyeqprec_{lex} \left< S_j' \right>_{1}^{nct(T)} = L'$
and hence $L \ll L'$. We thus proved that if two lists $T_i(S_n)$ and $T_i(S_n')$ of depth $p$ are such that $T_i(S_n) \ll
T_i(S_n')$ then
the sub-trees $S_n$ and $S_n'$ of depth $p$ which contain these lists are such that $S_n \curlyeqprec S_n'$ and thus that
the lists
$L$ and $L'$ of depth $p-1$ which contain $S_n$ and $S_n'$
are such that $L \ll L'$. It follows that $S$ and $S'$,
which are of depth $p-1$ and which contain $L$ and $L'$ are
such that $S \curlyeqprec S'$ and, by induction, that $C \curlyeqprec C'$.

Suppose now that C is canonical (and thus that C' is not).
Compare C and C' via a prefix traversal until we encounter two distinct sub-trees $S_n$ and $S_n'$.
As the list $L'$ which contains $S_n'$ is a permutation of the
list $L$ which contains $S_n$ and since $\forall j<n$, $S_j = S_j'$ then
$\exists m>n$, $S_m = S_n'$. As the list $L$ is sorted according to $\curlyeqprec$,
we have $S_n \curlyeqprec S_m$ and thus $S_n \curlyeqprec S_n'$. It follows that $C \curlyeqprec C'$. As the relation $C
\curlyeqprec C'$ is true
$\forall C' \in Iso(C)$, $C$ is $\curlyeqprec$-minimal over $Iso(C)$.

$\Rightarrow$ Now suppose that $C$ is $\curlyeqprec$-minimal over $Iso(C)$. Prove the contrapositive by assuming that $C$
is not canonical. 
Traverse $C$ as usual, and stop as soon as two sub-trees $S_n$ and $S_{n+1}$ are met such that $S_{n+1} \curlyeqprec S_n$. 
This necessarily happens since there exists at least a non sorted list of sub-trees because $C$ is not canonical.
Consider the tree $C'$ resulting from the permutation $\sigma$ which simply exchanges $S_n$ and $S_{n+1}$.
We have $C' \in Iso(C)$. As $S_{n+1} \curlyeqprec S_n$ then $\sigma(S_n) \curlyeqprec S_n$,  and it follows that $C'
\curlyeqprec C$
which contradicts the non canonicity hypothesis of $C$. $C$ is thus canonical.

\end{proof}

\section{Enumerating T-trees}

The rest of the study proposes on one hand a procedure allowing for the explicit production of only the canonical T-trees,
and on the other hand an algorithm to test and filter out non canonical T-trees. These two tools are meant to be integrated
as components within general purpose configurators, so as to avoid the exploration of solutions built on the basis of
redundant solutions of the inner structural problem of a given configuration problem.
We continue in the sequel  to call "configurations" the solutions of a structural problem . To generate a configuration
amounts to incrementally build a T-tree which satisfies all structural constraints.

\begin{definition}[Extension]
We call {\em extension} of a T-tree $C$, a T-tree $C'$ which results from adding nodes to $C$. 
We call {\em unit extension}, an extension which results from adding a single terminal node.
\end{definition}
The search space of a (structural) configuration problem can be described by a state graph $G=(V,E)$ where the nodes in $V$
correspond to valid (solution) T-trees and the edge $(t_1,t_2) \in E$ iff  $t_2$ is a unit extension of $t_1$. The goal of a constructive search procedure is to find a path in $G$ starting from the tree $(t,\left< \right>)$
(recall that $t$ is the type of the root object  in the configuration) and reaching a T-tree which respect all the problem
constraints (i.e. not only the constraints involved in the structural problem).

\begin{definition}[Canonical removal of a terminal node]
To canonically remove a terminal node from a T-tree $C$ not reduced to a single node consists in selecting its first non
empty T-list $T_i(C)$ (the first according to $\prec_{T_C}$) then to select a T-tree $C_j$ in this T-list : the first which
is not a leaf if one exists, or the last leaf otherwise. In the first case we recursively canonically remove one node of
$C_j$, in the other case, we simply remove the last leaf from the list.
\end{definition}

Notice that since the state graph is directed, the canonical removal of a leaf is not an applicable operation to a graph node (only unit extensions apply).
Canonical removal is technically useful to inductive proofs in the sequel.

\begin{proposition} \label{removal}
The canonical removal of a terminal node in a T-tree $C$ not reduced to a single node produces a T-tree  $C'$ such that
$C' \curlyeqprec C$.
\end{proposition}
\begin{proof}
Let $C_j$ be the $j^{th}$ T-tree of a T-list and $C_j'$ the tree resulting from the canonical removal of a node in $C_j$.
The proof is by induction over the depth $p$ of the root of $C_j$ in $C$. Let $L$ and $L'$ be the T-lists (of depth $p-1$)
containing $C_j$ and $C_j'$ : 
\begin{itemize}
\item if $C_j$ is a single node, it is removed from its T-list, thus $L' \ll L$.
\item else, if the canonical removal of a node of T-tree $C_j$ of depth $p$ produces a T-tree $C_j'$ such that $C_j'
\curlyeqprec C_j$ then
$\left<C_1, \ldots C_{j-1}, C_j', \ldots \right> \ll \left<C_1, \ldots C_{j-1}, C_j, \ldots \right>$ and thus $L' \ll L$.
\end{itemize}
In both cases, $L$ being the only T-list of $C$ modified to obtain  $L'$ (which transforms $C$ in $C'$), the same rationale
leads to $C' \curlyeqprec C$.

\end{proof}

\begin{proposition}\label{connexity}
Let $G$ be the state graph of a configuration problem. Its sub-graph $G_c$ corresponding to the only canonical T-trees is
connex.
\end{proposition}
\begin{proof}
It amounts to proving that any canonical T-tree can be reached by a sequence of canonical unit extensions from a T-tree
$(t,\left< \right>)$, or that (taken from the opposite side) the canonicity of a T-tree is preserved by canonical removal.
We proceed by induction over the height of T-trees.


\begin{itemize} 
\item Let $r$ be the depth of removed node. By definition of the canonical removal, it occurred at the end of its T-list,
which hence remains sorted after the change, and the parent T-tree (of depth $r-1$) remains canonical, since nothing else
is modified in the process. 
\item Now we show that whatever the value of $p$, if the canonical removal
of a node in a T-tree $C$ of depth $p$ preserves the canonicity of $C$, then the T-tree of depth $p-1$ which contains $C$
is remains canonical.
By the proposition \ref{removal}, the canonical removal of a node in a T-tree $C$ produces a T-tree $C'$ such that $C'
\curlyeqprec C$.
Canonical removal operates by selecting the first T-tree in a T-list that contains more than one node. 
If $C$ is not the last T-tree of its T-list, call $C_{right}$ the T-tree immediately after $C$ in the T-list. 
As $C' \curlyeqprec C$, we still have $C \curlyeqprec C_{right}$. 
If $C$ is not the first T-tree of its T-list, we call $C_{left}$ the T-tree immediately at the left of $C$ in the T-list.
As $C$ is the leftmost T-tree
containing more than a node, $C_{left}$ contains a single node, with the same root label as $C$ and $C'$. Since $C$
contained more than one node, $C'$ contains at least a node and $C_{left} \curlyeqprec C'$. 
Consequently, the canonical removal of a node in a T-tree (of depth $p$) of a T-list (of depth $p-1$) leaves the T-list
sorted. And the T-tree of depth $p-1$ which
contains this T-list, which is the only modified one, thus remains canonical.
\end{itemize}
We conclude that canonical removal preserves the canonicity of all the
sub-T-trees, whatever their depth in the T-tree. By this operation, a T-tree remains canonical. 
The sub-graph $G_c$ is thus connex.

\end{proof}

It immediately follows a practically very important corollary :

\begin{corollary}
A configuration generation procedure that filters out the interpretations containing a non canonical structural
configuration remains complete.

\end{corollary}
\begin{proof}
According to the proposition \ref{connexity}, to reject non canonical T-trees does not prevent to reach all canonical
T-trees, since each T-tree can be reached by a path sequence of canonical unit extensions from the empty T-tree.

\end{proof}
It thus suffices to add to any complete procedure enumeration of T-trees a canonicity test to obtain a procedure which
remains complete (in the set of equivalence classes for T-tree isomorphism) while avoiding the enumeration of isomorphic
(redundant) T-trees.

\section{Algorithms}

A test of canonicity straightforwardly follows from the definition of canonicity. It is defined by two functions :
\emph{Canonical} and \emph{Less} listed in pseudo code by the figure \ref{constraint}. 
We note $ct(T)$ the list of component types of $T$, sorted according to $\prec_{T_C}$, and by extension, as the labels of
nodes of a T-tree are types, we generalize these notations to $ct(C)$ for a given T-tree C.  Note that the function Less
compares T-trees with the same root type.

\begin{figure}[htb]
\vspace{-0.3cm}
  \begin{center}
    \begin{tabular}{|c|c|}
\hline
\hspace{0.2cm} \begin{minipage}{14cm}\small
\begin{tabbing}
\hspace*{0.3cm} \= \hspace*{0.3cm} \= \hspace*{0.3cm} \= \hspace*{0.3cm}
\= \hspace*{0.3cm} \= \hspace*{0.3cm} \= \kill
function {\bf Canonical$(C)$}\\
\{{\em returns True iff $C$ is canonical}\}\\
{\bf begin}\+\\
{\bf if} $C$ is a leaf {\bf then} {\bf return} True\\
Let $ct(C)=(T_1,\ldots,T_k)$\\
{\bf for } $i:=1$ {\bf to} $k$ {\bf do}\+\\
Let $(a_1,\ldots a_l)$ be the list $T_i(C)$\\
{\bf for} $j:=1$ {\bf to} $l$ {\bf do}\+\\
{\bf if} {\bf not}$({\rm Canonical}(a_j))$ {\bf then}\+\\
{\bf return} False\-\-\\
{\bf for} $j:=1$ {\bf to} $l-1$ {\bf do}\+\\
{\bf if} {\bf not}(Less$(a_j$, $a_{j+1})$)
{\bf then}\+\\
{\bf return} False\-\-\-\\
{\bf return}  True\-\\
{\bf end function}
\end{tabbing}
\end{minipage}
&
\hspace{0.2cm} \begin{minipage}{14cm}\small
\begin{tabbing}
\hspace*{0.3cm} \= \hspace*{0.3cm} \= \hspace*{0.3cm} \= \hspace*{0.3cm}
\= \hspace*{0.3cm} \= \hspace*{0.3cm} \= \hspace*{0.3cm} \= \kill
function {\bf Less$(C,C')$}\\
\{{\em Returns {\rm True} iff $C\curlyeqprec C'$}\}\\
{\bf begin}\+\\
{\bf if} $C$ is a leaf {\bf then} {\bf return} True\\
{\bf if} $C'$ is a leaf {\bf then} {\bf return} False\\
Let $ct(C)=(T_1,\ldots,T_k)$\\
{\bf for } $i:=1$ {\bf to} $k$ {\bf do}\+\\
Let $(a^i_1,\ldots,a^i_{l_a})$ be the list $T_i(C)$,\\
Let $(b^i_1,\ldots,b^i_{l_b})$ be the list $T_i(C')$\\
{\bf if} $(l_a<l_b)$ {\bf then return}  True\\
{\bf if} $(l_a>l_b)$ {\bf then return}  False\\
{\bf for} $j:=1$ {\bf to} $l_a$ {\bf do}\+\\
{\bf if} (Less$(a^i_j$, $b^i_j )=$False) {\bf then}\+\\
{\bf return}  False\-\-\-\\
{\bf return}  True\-\\
{\bf end function}
\end{tabbing}
\end{minipage}
\\
\hline
    \end{tabular}
    \caption{The functions \emph{Canonical} and \emph{Less}}
    \label{constraint}
  \end{center}
\vspace{-1cm}
\end{figure}

\vspace{-0,5cm}
\subsection{Complexity}

The worst case complexity of the function {\bf Less} is linear in $n$ ($\Theta(n)$), $n$ being the number of nodes of the
smallest T-tree. It is called at most once on each node. 
The function {\em Canonical} is of complexity $\Theta(n \log n)$ in the worst case.
It recursively calls itself for each sub-tree of its argument and tests that their T-lists are sorted via a call to {\bf
Less}.

\vspace{-0,2cm}
\subsection{Applications}
The algorithm described by the figure \ref{constraint} can be used as a constraint to filter out the non canonical
solutions of the structural sub-problem  of a configuration problem, and this is so whichever the enumeration procedure and
data structures are used (as possibly by example within the object oriented approach described in \cite{Mailharro98}). Il
can be integrated so that the test of canonicity is amortized over the search, if the T-tree corresponding to the currently
built configuration grows by unit extensions. In that case, the top part of the search made by "Canonical", that operates
on a T-tree that did not change, may be saved.

\section{Counting T-trees}

In this section, we show the potentially very important benefit that results from the enumeration of only the canonical
T-trees, compared with a standard exhaustive enumeration of all possible T-trees.
To this end, we count the total number of T-trees and of canonical T-trees  in a particular case of T-trees, those for
which each type (the label of nodes) may have children of a single type. 
The corresponding configuration problem  can be so defined : $p+1$ object types $T_0$, $T_1$, ... and $T_p$ that can be
inter connected by
the composition relations $R(T_0, T_1)$, $R(T_1, T_2)$, ... and $R(T_{p-1}, T_p)$. $T_0$ is the root type and there exists
exactly one object with this type. 
We may connect from 0 to $k$ objects of type $T_{i+1}$ to any object of type $T_i$. These T-trees are called $k$-connected.
We note $N_{p,k}$ (resp. $M_{p,k}$) the total number of $k$-connected T-trees (resp. canonical $k$-connected T-trees), of
maximal height $p$.

\subsection{Number of $k$-connected T-trees of depth p, $N_{p,k}$}
A T-tree of maximal height $p$ can be built by connecting from $0$ to $k$ T-trees of maximal height $p-1$ to a node root.
The number of arrangements of $i$ elements (some of which may be identical) among $N_{p-1,k}$ is
$\left(N_{p-1,k}\right)^i$. 
$N_{p,k}$ is thus recursively defined by : $N_{0,k} = 1$ (the tree containing a single root object root, thus no object of
$T_1$), $N_{1,k} = k+1$ (the configurations of 0 to $k$ objects of type $T_1$ without more children) and
$$\forall p>1, N_{p,k} = \sum_{i=0}^{i=k} \left(N_{p-1,k}\right)^i = \frac{\left(N_{p-1,k}\right)^{k+1}-1}{N_{p-1,k}-1}.$$
Then $N_{2,k}$ is in $\Theta(k^k)$ and $N_{p,k}$ is in $\Theta(k^{k^{p-1}})$.

\subsection{Number of canonical $k$-connected T-trees of depth p, $M_{p,k}$}

A canonical T-tree of maximal height $p$ can be obtained by connecting according to $\curlyeqprec$ from 0 to $k$ canonical
T-trees of maximal height $p-1$ to a root object.
The number of combinations of $i$ elements (some of which may be identical) among $M_{p-1,k}$ is
$\left(\stackrel{M_{p-1,k}+i-1}{i}\right)$.
$M_{p,k}$ is thus recursively defined by :
$M_{0,k} = 1$
 (the T-tree reduced to a single node) and
$$\forall p>0, M_{p,k} = \sum_{i=0}^{i=k} \left(\stackrel{M_{p-1,k}+i-1}{i}\right)
= \left(\stackrel{M_{p-1,k}+k}{k}\right) = \frac{(M_{p-1,k}+k)!}{M_{p-1,k}! k!}.$$ By the Stirling formula ($n! =
\sqrt{2\pi}.n^{n+\frac{1}{2}} e^{-n} + \epsilon(n)$), we get
$$M_{p,k} \simeq
\frac{1}{\sqrt{2\pi}}\frac{\left(M_{p-1,k}+k\right)^{\left(M_{p-1,k}+k+\frac{1}{2}\right)}}{\left(M_{p-1,k}\right)^{\left(M_{p-1,k}+\frac{1}{2}\right)}
k^{k+\frac{1}{2}}}.$$
$M_{1,k}=k+1$, $M_{2,k}$ is in $\Theta(4^k)$ and $M_{p,k}$ is in $\Theta(\frac{4^{k^{p-1}}}{k^{k^{p-2}}})$.
We see that $M_{p,k}$ is much smaller than $N_{p,k}$ for big values of $p$ and $k$. 
The table \ref{comparison} exhibits important benefits, even with very small values of $p$ and $k$. The case $p=2, k=2$
corresponds to the first 13 T-trees in figure \ref{fig:enumeration}. In the general case, where more than one composition
relation exists for each type, the impact of removing redundancies is even more important.
\begin{table}[htbp]
\vspace{-0.3cm}
\begin{center}
\begin{tabular}
{|c||c|c|c|c|}
\hline
$N_{p,k}$ / $M_{p,k}$ & $k=1$ & $k=2$ & $k=3$ & $k=4$\\
\hline
\hline
$p=1$ & 2 / 2 & 3 / 3 & 4 / 4 & 5 / 5\\
\hline
$p=2$ & 3 / 3 & 13 / 10 & 85 / 35 & 775 / 126\\
\hline
$p=3$ & 4 / 4 & 183 / 66 & 221436 / 8436 & 3.61 $10^{11}$ / 1.13 $10^7$\\
\hline
\end{tabular} 
\end{center} 
\caption{Comparison of $N_{p,k}$ and $M_{p,k}$ for small values of $p$ and $k$. For ($p=3$, $k=4$), we must have 4 objects of type
$T_1$, 16 objects of type $T_2$ and 64 objects of type $T_3$.}
\label{comparison}
\end{table} 


\vspace{-1,3cm}
\section{Conclusion}
Configuration problems are a difficult application of constraint programming, since they exhibit many isomorphisms. We have
shown that part of these isomorphisms, those stemming from the  properties of a sub-problem called the structural problem,
can be efficiently and totally tackled, by using low cost amortizable algorithm, so as to explore the only configurations
built upon a canonical solution of the structural sub-problem. We have also theoretically computed the numbers of canonical
and non canonical solutions of a simplified problem, showing that in this case already, there are much fewer canonical than
non canonical configurations.

These results extend the possibilities of dealing with isomorphisms in configurations, until today limited either to the
detection of the interchangeability of all yet unused individuals of each type or to the use of counters of non
configurable object counters (as in the ILOG software products\cite{Mailharro98}). Both approaches share the limitation of
not dealing with the structural bases of interchangeability (for example, in the case $14$ of the figure
\ref{fig:enumeration}, the two "B" are interchangeable, since they form the root of two equal trees, placed in the same
context (under the same "A"). The "D" which appear underneath are also interchangeable.

Our proposal allows to target in a near future the complete elimination of configuration isomorphisms, without needing
changes in the models (using counters by types rather than references to objects in relations).

\bibliographystyle{plain}
\bibliography{config}

\begin{thebibliography}{10}

\bibitem{AFM02}
Jérôme Amilhastre, Hélène Fargier, and Pierre Marquis.
\newblock Consistency restoration and explanations in dynamic
  csps----application to configuration.
\newblock {\em Artificial Intelligence}, 135(1-2):199--234, 2002.

\bibitem{AudHen01}
G.~Audemard and L.~Henocque.
\newblock The extended least number heuristic.
\newblock In Rajeev Gor{\'e}, Alexander Leitsch, and Tobias Nipkow, editors,
  {\em Proceedings of the First International Joint Conference, IJCAR, Sienne,
  Italie}, volume 2083 of {\em Lecture Notes in Computer Science}, pages
  427--442. Springer, June 2001.

\bibitem{barkerconnor89}
Virginia Barker, Dennis O'Connor, Judith Bachant, and Elliot Soloway.
\newblock Expert systems for configuration at digital: Xcon and beyond.
\newblock {\em Communications of the ACM}, 32:298--318, 1989.

\bibitem{Felfernig02}
Alexander Felfernig, Gerhard Friedrich, Dietmar Jannach, and Markus Zanker.
\newblock Semantic configuration web services in the cawicoms project.
\newblock In {\em Proceedings of the Configuration Workshop, 15th European
  Conference on Artificial Intelligence}, pages 82--88, Lyon, France, 2002.
\newblock http://www.cawicoms.org/.

\bibitem{FSB99}
Markus P.~J. Fromherz, Vijay~A. Saraswat, and Daniel~G. Bobrow.
\newblock Model-based computing: Developing flexible machine control software.
\newblock {\em Artificial Intelligence}, 114(1-2):157--202, October 1999.

\bibitem{JL87}
Joxan Jaffar and Jean~Louis Lassez.
\newblock Constraint logic programming.
\newblock In {\em in ACM Symposium on Principles of Programming Languages},
  pages 111--119, 1987.

\bibitem{JohnGeske99}
Ulrich John and Ulrich Geske.
\newblock Reconfiguration of technical products using conbacon.
\newblock In {\em Proceedings of AAAI'99-Workshop on Configuration}, pages
  48--53, Orlando, Florida, July 1999.

\bibitem{Mailharro98}
Daniel Mailharro.
\newblock A classification and constraint-based framework for configuration.
\newblock {\em AI in Engineering, Design and Manufacturing, (12)}, pages
  383--397, 1998.

\bibitem{McDermott82}
John~P. McDermott.
\newblock R1: A rule-based configurer of computer systems.
\newblock {\em Artificial Intelligence}, 19:39--88, 1982.

\bibitem{Mittal90AA}
Sanjay Mittal and Brian Falkenhainer.
\newblock Dynamic constraint satisfaction problems.
\newblock In {\em Proceedings of AAAI-90}, pages 25--32, Boston, MA, 1990.

\bibitem{Nareyek99}
Alexander Nareyek.
\newblock Structural constraint satisfaction.
\newblock In {\em Papers from the 1999 AAAI Workshop on Configuration,
  Technical Report, WS-99-0}, pages 76--82. AAAI Press, Menlo Park, California,
  1999.

\bibitem{Meyer99}
Harald~Meyer nauf'm Hofe.
\newblock Construct: Combining concept languages with a model of configuration
  processes.
\newblock In {\em Papers from the 1999 AAAI Workshop on Configuration,
  Technical Report, WS-99-0}, pages 17--22, 1999.

\bibitem{Plain2002}
Kevin~R. Plain.
\newblock Optimal configuration of logically partitionned computer products.
\newblock In {\em Proceedings of the Configuration Workshop, 15th European
  Conference on Artificial Intelligence}, pages 33--34, Lyon, France, 2002.

\bibitem{sabinfreudercomposite96}
Daniel Sabin and Eugene~C. Freuder.
\newblock Composite constraint satisfaction.
\newblock In {\em Artificial Intelligence and Manufacturing Research Planning
  Workshop}, pages 153--161, 1996.

\bibitem{soininen99fixpoint}
Timo Soininen, Esther Gelle, and Ilkka Niemela.
\newblock A fixpoint definition of dynamic constraint satisfaction.
\newblock In {\em Proceedings of CP'99}, pages 419--433, 1999.

\bibitem{soininen01representing}
Timo Soininen, Ilkka Niemela, Juha Tiihonen, and Reijo Sulonen.
\newblock Representing configuration knowledge with weight constraint rules.
\newblock In {\em Proceedings of the {AAAI} Spring Symp. on Answer Set
  Programming: Towards Efficient and Scalable Knowledge}, pages 195--201, March
  2001.

\bibitem{Tiihonen02}
Juha Tiihonen, Timo Soininen, Ilkka Niemela, and Reijo Sulonen.
\newblock Empirical testing of a weight constraint rule based configurator.
\newblock In {\em Proceedings of the Configuration Workshop, 15th European
  Conference on Artificial Intelligence}, pages 17--22, Lyon, France, 2002.

\bibitem{Weigel1998}
Rainer Weigel, Boi Faltings, and Marc Torrens.
\newblock Interchangeability for case adaptation in configuration problems.
\newblock In {\em Workshop on Case-Based Reasoning Integrations (AAAI-98)},
  volume Technical Report WS-98-15, pages 166--171, Madison, Wisconsin, USA,
  July 1998. AAAI Press.

\bibitem{Ylinen02}
Katariina Ylinen, Tomi Männistö, and Timo Soininen.
\newblock Configuring software products with traditional methods - case linux
  familiar.
\newblock In {\em Proceedings of the Configuration Workshop, 15th European
  Conference on Artificial Intelligence}, pages 5--10, Lyon, France, 2002.

\end{thebibliography}

\end{document}